\documentclass{article}

\usepackage{graphicx}
\usepackage{lscape}

\title{A Review on Learning Planning Action Models for Socio-Communicative HRI}

\author{Ankuj Arora, Humbert Fiorino, Damien Pellier and Sylvie Pesty\\[2mm]
Laboratoire LIG, Universit\'e Grenoble-Alpes, \\Grenoble, France}

\date{June 2016}

\begin{document}

\maketitle

\begin{abstract}
For social robots to be brought more into widespread use in the fields of companionship, care taking and domestic help, they must be capable of demonstrating social intelligence.
%This intelligence opens the application possibilities of social robots in the roles of companionship, care taking and domestic help; but also raises several problems of psychosocial acceptability.
In order to be acceptable,
%and be able to subtly blend into their physio-social context,
they must exhibit socio-communicative skills. Classic approaches to program HRI from observed human-human interactions fails to capture the subtlety of multimodal interactions as well as the key structural differences between robots and humans. The former arises due to a difficulty in quantifying and coding multimodal behaviours, while the latter due to a difference of the degrees of liberty between a robot and a human. However, the notion of reverse engineering from multimodal HRI traces to learn the underlying behavioral blueprint of the robot given multimodal traces seems an option worth exploring. With this spirit, the entire HRI can be seen as a sequence of exchanges of speech acts between the robot and human, each  act treated as an action, bearing in mind that the entire sequence is goal-driven. Thus, this entire interaction can be treated as a sequence of actions propelling the interaction from its initial to goal state, also known as a plan in the domain of AI planning. In the same domain, this action sequence that stems from plan execution can be represented as a trace. AI techniques, such as machine learning, can be used to learn behavioral models (also known as symbolic action models in AI), intended to be reusable for AI planning, from the aforementioned multimodal traces. This article reviews recent machine learning techniques for learning planning action models which can be applied to the field of HRI with the intent of rendering robots as socio-communicative.\\
\end{abstract}

\section{Introduction}

With the near simultaneous advances in mechatronics on the engineering side and ergonomics on the human factors side, the field of social robotics has seen a significant spike in interest in the recent years. Driven with the objective of rendering robots as socio-communicative, there has been an equally heightened interest towards researching techniques to endow robots with cognitive, emotional and social skills. The strategy to do so draws inspiration from study of human behaviors. For robots, social and emotive qualities not only lubricate the interface between humans and robots, but also promote learning, decision making and so on. These qualities strengthen the possibility of acceptability and emotional attachment to the robot \cite{breazeal2003emotion,breazeal2004social}. This acceptance is only likely if the robot fulfils a fundamental expectation that one being has of the other: not only to do the right thing, but also at the right time and in the right manner \cite{breazeal2003emotion}. This social intelligence or 'commonsense' of the robot is what eventually determines its social acceptability in the long run.

Commonsense, however,  is not that common. Robots can, thus, only learn to be acceptable with experience. However, teaching a humanoid the subtleties of a social interaction is not evident. Even a standard dialogue exchange integrates the widest possible panel of signs which intervene in the communication and are difficult to codify (synchronization between the expression of the body, the face, the tone of the voice, etc.). In such a scenario, learning the behavioral model of the robot is a promising approach.
%The most common strategy used to render robots as socio-communicative is to analyze, model and implement human behaviors either by observation, imitation or demonstration \cite{argall2009survey}. The limitation with this approach is the scaling of the human model to the interaction capabilities of the robots in terms of physical limitations (degrees of freedom) as well as perception, action and reasoning.

Thus, another way of solving this problem is given a set of HRI traces, to learn the interaction script or the behavioral model of the robot which governs this interaction. This learning can be conducted with the help of some fairly recent and other ongoing advances in the field of AI. In the field of AI planning for instance, this entire interaction can be viewed as a series of actions (where each speech act is treated as an action) which take the system from the initial to the goal state, the goal being the successful termination of the interaction \cite{perrault1980plan}.  In the literature, AI (or Automated) planning has been used in HRI for robot action planning and reasoning \cite{alami2005task} . There has been little work done in using AI planning approaches to empower robots with socio-communicative abilities.
%However, even this standard dialogue exchange integrates the widest possible panel of signs which intervene in the communication and are difficult to codify (synchronization between the expression of the body, the face, the tone of the voice, etc.). In such a scenario, learning of behavioral (action) models is the most plausible approach.

In such domains, planners now leverage recent advancements in machine learning (ML) to recreate the blueprint of the actions applicable to the domain, but cannot be easily identified or programmed, alongwith their signatures, preconditions and effects. Several classical and recent ML techniques can be leveraged to reproduce the underlying behavioral model. This model, once learnt, can further render the humanoid as autonomous and pioneer future HRI interactions. This is a conscious and directed effort to decrease laborious manual coding and increase quality. This article briefly reviews recent machine learning techniques for learning planning action models for the field of HRI.

This article is organized as follows: we start by briefly introducing and explaining the interplay between Automated Planning (AP), Machine Learning (ML) and HRI to solve a common problem. We then represent a classification of various techniques of learning action models, citing examples of each. These examples are then detailed in the following section. We then briefly discusses the persisting issues in the field despite the advances, and terminate with a conclusion.
\section{Problem Statement}

\begin{figure}[h]
    \centering
    \includegraphics[width=60mm]{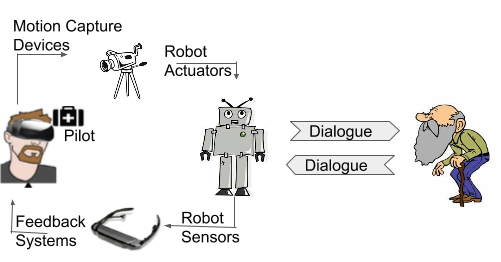}
    \caption{Step 1 - Initial Run of HRI experiment by beaming \cite{bailly2015beaming}}
    \label{fig:planningLearning1}
    \end{figure}

  Consider the case where Automated Planning (AP) were to be used to construct the core behavioral model of a robot to govern its multimodal interactions with humans. It is very difficult, if not impossible, to fine tune such a model by a domain expert to inculcate and account for subtle human behaviours which arise and interplay even in the simplest dialog exchange. However, using ML techniques, it is possible to learn this model from an actual HRI. As demonstrated in the figure~\ref{fig:planningLearning1}, the HRI can be viewed as a planning problem: in an initial run of the interaction, the robot speech acts are governed by observation, imitation or demonstration techniques \cite{argall2009survey,verstaevel2015principles}.  One particular approach that seems promising is that of 'beaming' by human pilots \cite{bailly2015beaming}. Thanks to this technique, a human operator can perceive, analyze, act and interact with a remote person through robotic embodiment. A human operator will solve both the scaling and social problems by optimally exploiting the robots affordances. The following speech act exchange sequence between the robot and human is treated as a sequence of actions which constitutes the trace set (also called the execution experience). This speech act exchange is what drives the interaction from its initial to goal state. An initial state or a goal is composed of a set of predicates. A predicate is a set of constant symbols or variable symbols. It is a set of constant symbols in case it is grounded, in which case it evaluates to true or false. Thus, a single trace is constituted of: initial state predicates, speech act sequence, and the final state predicates.

\begin{figure}[h]
    \centering
    \includegraphics[width=60mm]{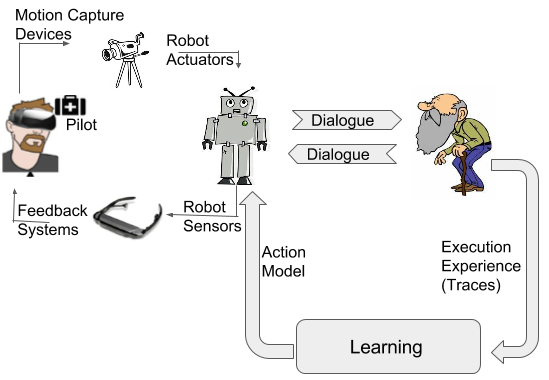}
    \caption{Step 2 - Learning the behavioral model from collected traces }
    \label{fig:planningLearning2}
    \end{figure}

These traces are then fed to the learner (see figure ~\ref{fig:planningLearning2}), whose role is to learn the behavioral (action) model $m$ that serves as the 'blueprint' of the actions. An action model is defined by $(a, Pre, Add, Del)$, where $a$ is an action name with zero or more variables (as parameters), $Pre$, $Add$ and $Del$ being the precondition list, add list and delete list, respectively. The precondition list is the set of conditions (eventually a conjunction of predicates) which need to be satisfied for the action to be triggered in a particular state. The add and delete lists are the set of grounded predicates which will be added or deleted respectively from the current state, upon the application of the action to the current state. This action application then produces the next state, which upon successive action applications leads to the goal state. In the field of AI planning, the model is represented in a standard language called the Planning Domain Definition Language (PDDL). It has been the official language for the representation of the problems and solutions in all of the International Planning Competitions (IPC) which have been held 1998 onwards \cite{mcdermott1998pddl}.

    \begin{figure}[h]
    \centering
    \includegraphics[width=80mm]{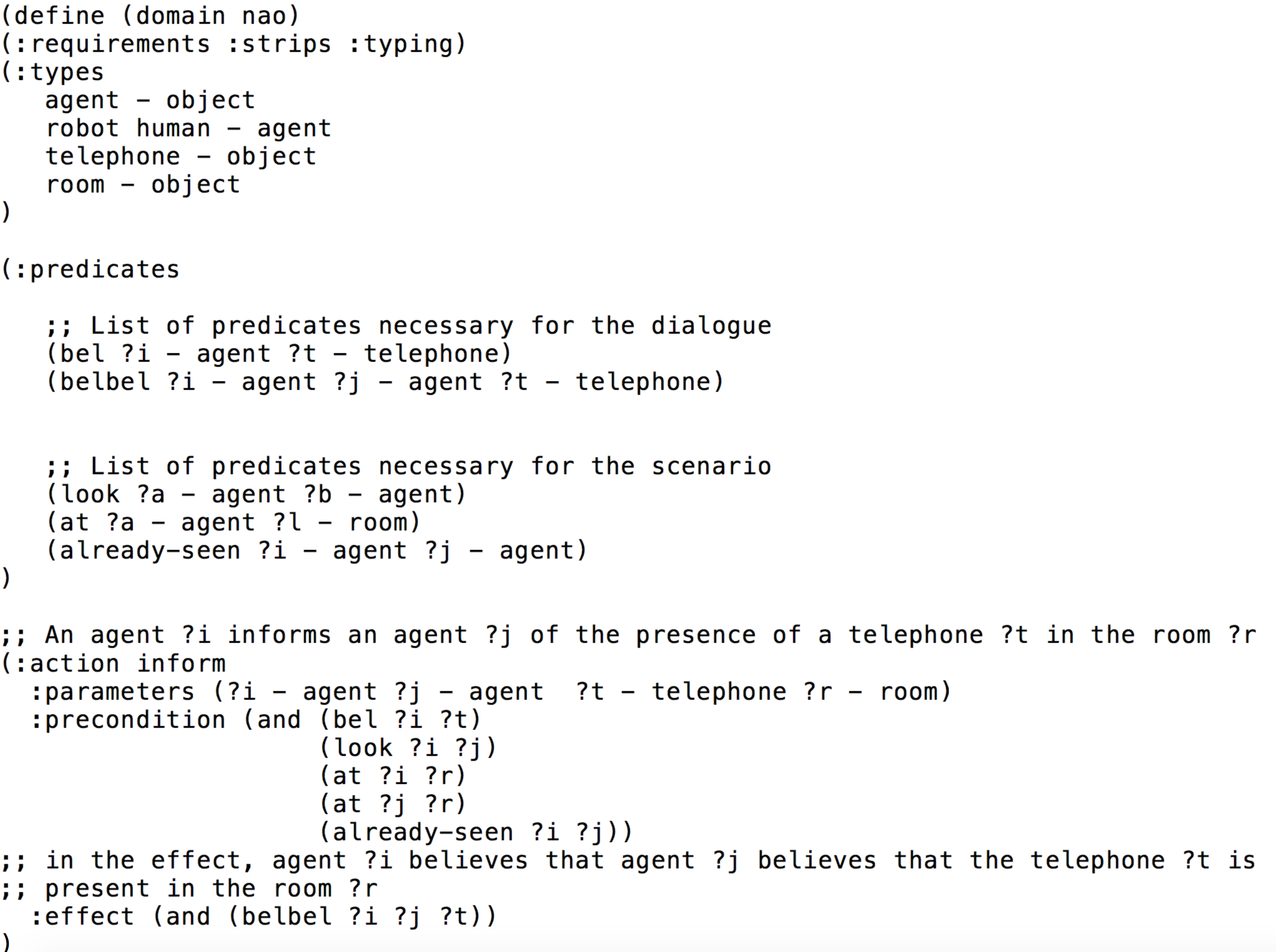}
    \caption{Domain Description and Schema for Operator 'inform' in an HRI domain }
    \label{fig:actionModelSchema}
    \end{figure}

%The constituents of PDDL are defined as follows \cite{mcdermott1998pddl}:

%\begin{itemize}
%\item Domain Description: This is constituted by the domain name definition, requirement definition (declaration of used model-elements to the planner), object-type hierarchy definition (name type definition), definition of predicates (blueprint for symbolic facts), and an operator schema definition (operator blueprints which must be instantiated during implementation).

%\item The problem description consisted of a problem-name definition, associated domain name definition, object definitions, initial and goal state definitions of the system.
%\end{itemize}

A sample action called 'inform' in the PDDL language is represented in the figure~\ref{fig:actionModelSchema}. The objective of this action is for the agent (in this case the robot) to inform a human about the presence of a telephone in the room. The preconditions for this action: both the robot and human are in the room, the robot believes in the presence of a telephone in the room, and that the robot has seen the human; are represented in the form of predicates. As an effect of this action, the robot believes that the human believes in the presence of a telephone in the room, signifying its successful execution of the action 'inform'.

    \begin{figure}[h]
    \centering
    \includegraphics[width=70mm]{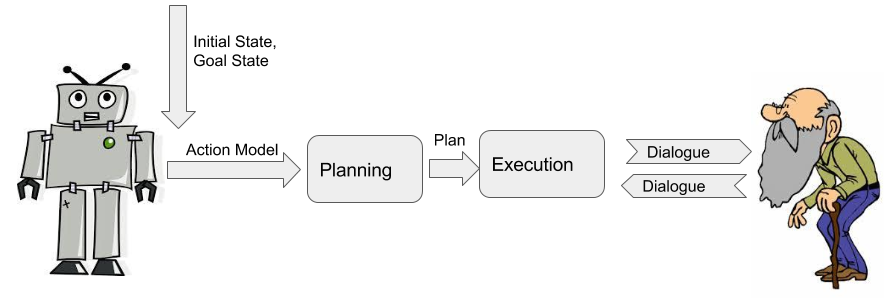}
    \caption{Step 3 - Autonomous re-run of HRI}
    \label{fig:planningLearningRerun}
    \end{figure}

    The challenge lies scripting this action model with more complex speech acts, relying solely on the expertise of a domain expert who also in his own right, is likely to commit errors while scripting.

The effort required by the domain expert to script this subtle and delicate humanoid behavioral model can be diminished by ML. The work done in ML goes hand in hand with the long history of planning, as ML is viewed as a potentially powerful means of endowing an agent with greater autonomy and flexibility. It often compensates for the designer's incomplete knowledge of the world that the agent will face. The developed ML techniques could be applied across a wide variety of domains to speed up planning. This is done by learning the underlying behavioral model from the experience accumulated during the planning and execution phases (refers to speech act exchanges). These employed learning techniques vary widely in terms of context of application, technique of application, adopted learning methodology and information learned.

Once the model $m$ is learnt, it is fed to a planner (see figure ~\ref{fig:planningLearningRerun}) along with an initial state $s0$ and goal state $g$. Together, all three constitute a planning problem which is defined by $(s0, g, m)$.  A solution to a planning problem is a plan composed of an action sequence $(a1 , a2 , an )$, where the actions guide the transition of the system from the initial to the goal state \cite{zhuo2013action}. Thus, the learnt model can be re-usable to plan future dialogue sequences between the robot and the human, in such a way that the need of a 'teacher' to govern the robot behavior is suppressed, and the robot can interact autonomously.

In summary, Machine Learning (ML) is increasingly being used to resolve the aforementioned planning problem. This article tries to classify various approaches based on several criterion.
%It then broadly highlights some persisting open issues with the discussed approaches.

\section{Learning Planning Action Models}

\begin{figure}[t]
    \centering
    \includegraphics[width=90mm]{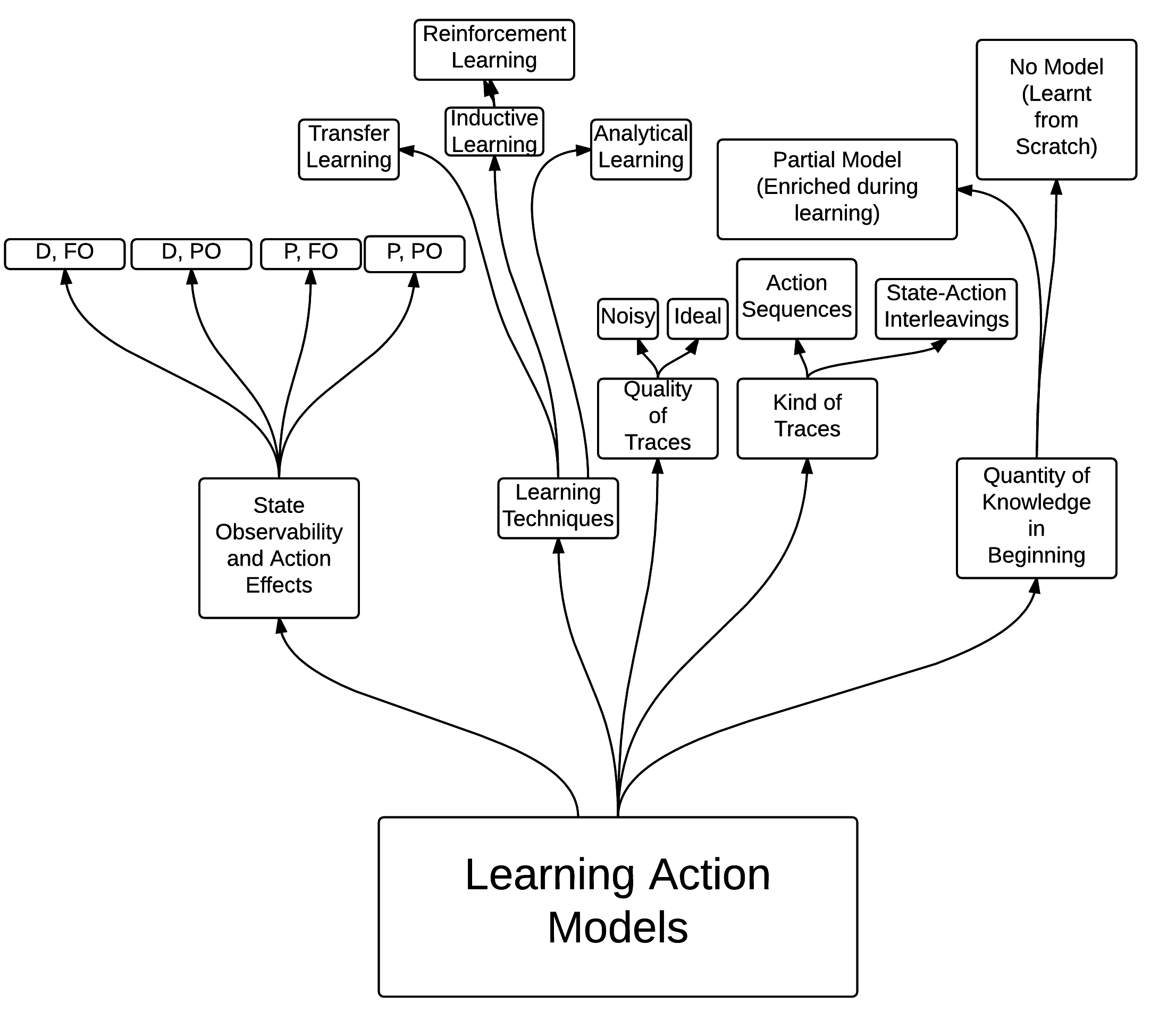}
    \caption{Learning Planning Action Models (D=Deterministic, P=Probabilistic, FO=Fully Observable, PO=Partially Observable)}
    \label{fig:planningLearningModel}
    \end{figure}

The techniques for learning planning action models can be classified as depicted in figure~\ref{fig:planningLearningModel}.

        \subsection{State Observability and Action Effects}
        The determination of the current state of the system after the action execution may be flawed because of a faulty sensor calibration. Thus, in the case of partial observability of a system, it may be assumed to be in one of a set of 'belief states'.

        Similarly action effects may be probabilistic, which means that in a real world scenario, it is not necessary that a unitary action be applicable to a state. On the contrary, multiple actions may be applied, each with a different execution probability.

        Keeping these variations of action effects and state observability in mind, we define four categories of implementations:
        \begin{itemize}
            \item Deterministic effects, full state observability: For example, the EXPO \cite{wang1996learning,jimenez2013integrating,gil1992acquiring} system.
            %refined incomplete planning operators, that is,  operators with some over-general preconditions and missing effects by means of ORM (operator refinement method). EXPO does this by generating plans and monitoring their execution to detect the differences between the state predicted according to the internal action model and the observed state. EXPO then constructs a set of specific hypotheses to fix the detected differences. After being heuristically filtered, each hypothesis is tested in turn with an experiment and a plan is constructed to achieve the situation required to carry out the experiment. This approach is also interesting because it also allows for the formation of micro operators in situations where only some effects of operator are required.

            \item Deterministic effects, partial state observability: In this family, the system may be in one of a set of 'belief states' after the execution of each action. For example, ARMS (Action-Relation Modelling System) \cite{yang2007learning}.

            \item Probabilistic effects, full state observability:  For example, PELA (Planning, Execution and Learning Architecture)\cite{jimenez2008architecture}.
%            , as its name suggests, performs three principal functions: (i) Planning the actions that solve a given problem (2) Execution of plans and classification of the execution outcomes. PELA executes plans one action at a time in the environment and labels the actions executions according to their outcomes as success, failure or dead-end. The learning component allows PELA to generate probabilistic rules about the execution of actions. PELA generates these rules from the execution of plans and compiles them to upgrade its deterministic planning model. This is done by performing multiclass classification, which further consists of finding the smallest decision tree that fits a given data set. The common way to find these decision trees is following a Top-Down Induction of Decision Trees (TDIDT) algorithm \cite{quinlan1986induction}. This approach builds the decision tree by splitting the learning examples according to the values of a selected attribute that minimize a measure of variance along the prediction variable.

            \item Probabilistic Effects, Partial State Observability: Barring a few initial works in this area, this classification remains as the most understudied one to date, with no general approach in sight either (for example, Yoon et al. \cite{yoon2007towards}).

        \end{itemize}

\subsection{Learning Techniques}
This section introduces some classic as well as recently prominent learning techniques that have been successfully used in learning action models. The following subsections have not been conceptualized as learning families, but as orthogonal (and sometimes overlapping) techniques.

\subsubsection{Inductive and Analytical Learning}
\begin{itemize}
\item Inductive learning: The learning system is confronted with a hypothesis space $H$ and a set of training examples $D$. The desired output is a hypothesis $h$ from $H$ that is consistent with these training examples. Inductive methods generate statistically justified hypotheses \cite{zimmerman2003learning}. The heart of the learning problem is generalizing successfully from examples. In these cases, inductive techniques that can identify patterns over many examples in the absence of a domain model can come in handy. One prominent inductive learning technique is that of decision tree and regression tree learning. Regression trees offer the advantages of being able to predict a continuous variable and the ability to model noise in the data. A regression tree predicts a value along the dependent dimension for all environmental observations, in contrast to a decision tree, which enables a prediction along a categorical variable (i.e., class).

\item Analytic learning: The learning system is confronted with the same hypothesis space and training examples as for inductive learning. However, the learner has an additional input: background knowledge $B$ that can explain observed training examples. The desired output is a hypothesis $h$ from $H$ that is consistent with both the training examples $D$ and the background knowledge $B$ \cite{zimmerman2003learning}. Analytic learning leans on the learner's background knowledge to analyze a given training instance to identify the relevant features.
\end{itemize}
More details about classical techniques which have been comprehensively used in operator learning can be found in \cite{zimmerman2003learning}. The current article sheds light on certain interesting techniques which have more recently come to light and offer interesting possibilities with respect to the task at hand, which is that of learning operators.
\subsubsection{Transfer Learning}

Many machine learning methods work well only under a common assumption: the training and test data are drawn from the same feature space and the same distribution. When the distribution changes, most statistical models need to be rebuilt from scratch using newly collected training data. In many real world applications, it is expensive or impossible to re-collect the needed training data and rebuild the models. It would be nice to reduce the need and effort to re-collect the training data. In such cases, knowledge transfer or transfer learning between task domains would be desirable. Transfer learning \cite{pan2010survey}, in contrast, allows the domains, tasks, and distributions used in training and testing to be different. In the real world, we observe many examples of transfer learning. For example, we may find that learning to recognize apples might help to recognize oranges. Transfer learning aims to extract the knowledge from one or more source tasks and applies the knowledge to a target task when the latter has fewer high-quality training data \cite{pan2010survey}.

%In transfer learning, we have the following three main research issues: (1) What to transfer; (2) How to transfer; (3) When to transfer. “What to transfer” asks which part of knowledge can be transferred across domains or tasks. Some knowledge is specific for individual domains or tasks, and some knowledge may be common between different domains such that they may help improve performance for the target domain or task. After discovering which knowledge can be transferred, learning algorithms need to be developed to transfer the knowledge, which corresponds to the “how to transfer” issue. “When to transfer” asks in which situations, transferring skills should be done. Likewise, we are interested in knowing in which situations, knowledge should not be transferred. In some situations, when the source domain and target domain are not related to each other, brute-force transfer may be unsuccessful. In the worst case, it may even hurt the performance of learning in the target domain, a situation which is often referred to as negative transfer.

The advantages of using transfer learning are centered around the fact that a change of features, domains, tasks, and distributions from the training to the testing phase does not require the statistical model to be rebuilt. The disadvantages, however, are listed as follows:
\begin{itemize}

\item Many proposed transfer learning algorithms assume that the source and target domains are related to each other in some sense. However, if the assumption does not hold, negative transfer may happen, which is worse than no transfer at all (for example, an American tourist learning to drive on the left side of the road in the UK for the first time). In order to avoid negative transfer learning, we need to first study transferability between source domains and target domains. Based on suitable transferability measures, we can then select relevant source domains/tasks to extract knowledge for learning the target tasks.

\item Most existing transfer learning algorithms so far have focused on improving generalization across different distributions between source and target domains or tasks. In doing so, they assumed that the feature spaces between the source and target domains are the same. However, in many applications, we may wish to transfer knowledge across domains or tasks that have different feature spaces, and transfer from multiple such source domains. This type of transfer learning is referred to as heterogeneous transfer learning, which is a persisting challenge.

\item Has mainly been applied to small scale applications \cite{pan2010survey}.

\end{itemize}

One particular implementation is an algorithm called LAWS (Learn Action models with transferring knowledge from a related source domain via Web Search) \cite{zhuo2011cross}.
%Given just a limited amount of training data, it make use of action-models already created beforehand in other related domains, which are called source domains, to help learn actions in a target domain. The target domain and a related source domain are bridged by searching Web pages related to the target domain and the source domain, and then building a mapping between them by means of a similarity function done by calculating the similarity between their corresponding Web pages.  The similarity is calculated using the Kullback-Leibler (KL) divergence or Maximum Mean Discrepancy (MMD). Based on the calculated similarity, a set of weighted constraints, called web constraints, are built. Based any available example plan traces in the target domain, other constraints such as state constraints, action constraints and plan constraints, are also built. All the constraints (web/state/action/plan constraints) are solved using a weighted MAX-SAT solver, and target-domain action models are generated based on the solution to the constraint satisfaction problem.

\subsubsection{Reinforcement Learning}
Reinforcement learning (RL) \cite{barto1992reinforcement} is a specific case of inductive learning, and defined more clearly by characterizing a learning problem instead of a learning technique. A general reinforcement learning problem can be seen as composed of just three elements: (1) goals an agent must achieve, (2) an observable environment, and (3) actions an agent can take to affect the environment. Through trial-and-error online visitation of states in its environment, such a reinforcement learning system seeks to find an optimal policy for achieving the problem goals. The strength of reinforcement learning lies in its ability to handle stochastic environments in which the domain theory is either unknown or incomplete. With respect to the planning-learning goal dimension, reinforcement learning can be viewed as both 'improving plan quality' (the process moves toward the optimal policy) and 'learning the domain theory' (begins without a model of transition probability between states) \cite{zimmerman2003learning}. However, one of the major drawbacks of RL stems from the fact that in its bid to achieve particular goals, it cannot gather general knowledge of the system dynamics, leading to a problem of generalization. RL is particularly interesting for robotics, for this approach often involves learning to achieve particular goals, without gathering any general knowledge of the world dynamics. As a result, the robots can learn to do particular tasks without having trouble generalizing to new ones \cite{pasula2004learning}. For example, LOPE (Learning by Observation in Planning Environments) \cite{garcia2000integrated}.

\subsubsection{Surprise-Based Learning(SBL)}
%In Surprise-Based Learning (SBL) \cite{ranasinghe2008surprise} a surprise is produced if the latest prediction is noticeably different from the latest observation. The algorithm must not only detect a surprise, it must also distinguish a possible cause for the surprise by investigating the change in features (see figure ~\ref{fig:sbl}). After performing an action, the world is sensed via the perceptor module which extracts feature information from one or more sensors. If the algorithm had made a prediction, the surprise analyzer will validate it. If the prediction was incorrect, the model modifier will adjust the world model accordingly. Based on the updated model the action selector will perform the next action so as to repeat the learning cycle \cite{ranasinghe2008surprise}.

\begin{figure}[h]
    \centering
    \includegraphics[width=70mm]{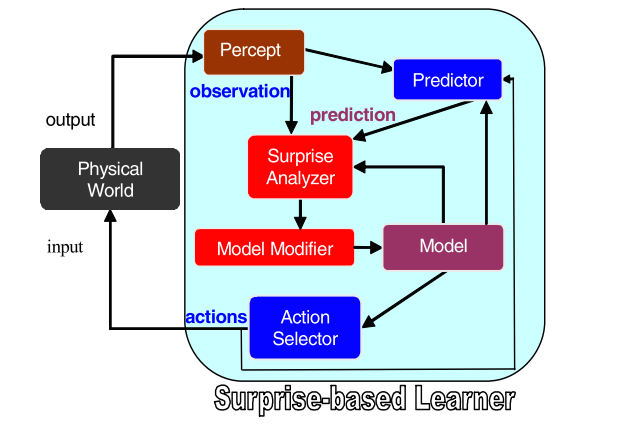}
    \caption{Overview of surprise based learning \cite{ranasinghe2008surprise}}
    \label{fig:sbl}
    \end{figure}

In Surprise-Based Learning (SBL) \cite{ranasinghe2008surprise} a surprise is produced if the latest prediction is noticeably different from the latest observation. After performing an action, the world is sensed via the perceptor module which extracts feature information from one or more sensors. If the algorithm had made a prediction, the surprise analyzer will validate it. If the prediction was incorrect, the model modifier will adjust the world model accordingly. Based on the updated model the action selector will perform the next action so as to repeat the learning cycle (see figure~\ref{fig:sbl}).

A series of approaches based on SBL have used Goal Driven Autonomy (GDA). GDA is a conceptual model for creating an autonomous agent that monitors a set of expectations during plan execution, detects when discrepancies occur, builds explanations for the cause of failures, and formulates new goals to pursue when planning failures arise. In order to identify when planning failures occur, a GDA agent requires the planning component to generate an expectation of world state after executing each action in the execution environment. The GDA model thus provides a framework for creating agents capable of responding to unanticipated failures during plan execution in complex, dynamic environment \cite{weber2012learning}. For example, the system FOOLMETWICE \cite{molineaux2014learning}.

        \subsection{Quality of traces} The execution traces may be classified into pure or adulterated as follows:

        \begin{itemize}
            \item Noisy: The traces can be adulterated because of sensor miscalibration or faulty annotation by a domain expert. For instance, AMAN (Action-Model Acquisition from Noisy plan traces) \cite{zhuo2013action} belongs to this category.
            %finds a domain model that best explains the observed noisy plan traces. First, a set of candidate domain models is built by scanning each action in plan traces and substitute its instantiated parameters with their corresponding variables , thus building a set of action schemas. This is followed by a graphical model to capture the relationship between the current state, correct action, observed action and the domain model. Afterwards the parameters of the graphical model are learnt, following which AMAN generates a set of action models according to the learnt parameters. Specifically, first the observed noisy plan traces are exploited to predict correct plan traces and the domain model based on the graphical model. Then the correct plan traces are executed to calculate the reward of the predicted correct plan traces according to a predefined reward function.  The AMAN algorithm framework belongs to a family of policy gradient algorithms, which have been successfully applied to complex problems.
            \item Ideal: There is no discrepancy between the ideal action and the recorded action. For example, the system OBSERVER \cite{wang1996learning} falls into this category.
        \end{itemize}
        \subsection{Kind of traces} This refers to the elements (state information, action information) which constitute the traces. These can be divided into the following:
        \begin{itemize}
            \item Action Sequences: Refers to the case where the executed plan traces can be represented in the form of a sequence of action executions.
            For example, Opmaker \cite{mccluskey2002interactive}.
            %is a mixed initiative knowledge acquisition tool for inducing parameterized, hierarchical operator descriptions from example action sequences and declarative domain knowledge, with the minimum of user interaction. It uses a high level language called OCL (Object Centered Language) for domain modelling. As Opmaker creates an operator schema from each action in a training solution sequence, it asks the user to input, if needed, the target state that each object would occupy after execution of the action. The algorithm induces operators, using a high level, partial model of the domain, as well as using the causal structure of the example sequence to track the state of an object.  It is implemented inside of a graphic tool called GIPO (Graphical Interface for Planning with Objects), which facilitates  domain knowledge capture and  domain modelling (\cite{mccluskey2002interactive,jilani2014automated}).

        \item State-Action Interleavings: The case where the executed plan traces can be represented in the form of a sequence of alternate state-action representations. For example, LAMP (Learning Action Models from Plan traces) proposed by \cite{zhuo2010learning}.
        %learn action models with quantifiers and logical implications . Firstly,  the input plan traces(including observed states and actions) are encoded into propositional formulas, which is a conjunction of ground literals to store into a database as a collection of facts. These formulas are very close to PDDL in terms of representation. Secondly, candidate formulas are generated according to the predicate lists and domain constraints. Thirdly, a Markov Logic Network (MLN) uses the formulas generated in the above two steps to learn the corresponding weights of formulas to select the most likely subset from the set of candidate formulas. Finally, this subset is converted into the final action models \cite{zhuo2010learning}.
        \end{itemize}

    \subsection{Availability of model in beginning} Before the learning phase begins, the action model may exist in one of the following capacities:
    \begin{itemize}
        \item No Model: This refers to the fact that the no information on the actions that constitute the model is available in the beginning, and the entire model must be learnt from scratch.
        For example, OBSERVER \cite{wang1996learning}.

%        For example, OBSERVER is a system that incrementally learns planning operators by first observing expert agents performing tasks in the domain, and then attempting to form plans of its own from its learned knowledge. OBSERVER  allows: (i) planning with incomplete and incorrect operators (ii) plan repair upon execution failure, and (iii) integrating planning, learning, and execution. OBSERVER generates a  specific representation then a general one using an algo similar to version spaces \cite{wang1996learning}

%        For example, OBSERVER (\cite{wang1996learning}) is a system that integrates planning, learning, and execution. It's learning module learns the preconditions and effects of operators by observing expert solution traces and to further refine the operators through practice by solving problems in the simulator using a learning-by-doing mechanism. Learning the operator preconditions consists of creating and updating both most specific representation and a most general representation for the preconditions, based on operator executions while solving practice problems. It also learns operator effects by generalizing the delta-state (the difference between post-state and pre-state) from multiple observations.
        \item Partial Model: Some elements of the model are available to the learner in the beginning, and the model is enriched with more knowledge at the end of the learning phase. For example, RIM (Refining Incomplete planning domain Models through plan traces) \cite{zhuo2013refining}.
%        begins with the step of collecting sets of predicates, action schemas and macro-operator schemas from incomplete action models  and plan traces . In the following steps, it constructs sets of soft and hard constraints to ensure that the learned domain model can best explain the input plan traces and incomplete action models. These constraints are designed to be soft, directing the MAX-SAT algorithm in the learning step towards learning the most probable complete description of actions and macro-operators. Then, a set of hard constraints is enforced that must be satisfied by action models and macro-operators. Finally, these constraints are solved using a weighted MAX-SAT solver to obtain sets of macro-operators and (refined) action models. Macro-operators are used in this context to increase the accuracy of the incomplete model.
        %it as input a partially specified domain model (with missing preconditions and effects in the actions), as well as a set of plan traces that are known to be correct. It outputs a 'refined' model that not only captures additional precondition/effect knowlege about the given actions, but also 'macro actions'.  We convert the set of plan traces to a sequence database in order to make use of the frequent pattern mining algorithm for extracting macro-operators. The macro-operators increases the accuracy of the incomplete model. The MAX-SAT framework for learning, where the constraints are derived from the executability of the given plan traces, as well as the preconditions/effects of the given incomplete model.

        \end{itemize}

\subsection{Representation Language}
The ideal language would be able to compactly model every action effect the agent might encounter, and no others. Choosing a good representation language provides a strong bias for any algorithm that will learn models in that language. Some languages and their features are summarized in the table~\ref{Language}.
%Some representation languages include: (i) PDDL \cite{mcdermott1998pddl} (ii) STRIPS (STanford Institute Problem Solver) language: Sub-language of PDDL \cite{fikes1971strips} (iii) OCL (Object Centered Language): Representation centered around objects instead of states \cite{mccluskey2002interactive} (iv) STRIPS+WS: STRIPS with the facility to introduce functional terms which model object creation \cite{walsh2008efficient}.

\begin{table}[]
\centering
\caption{Representation Languages}
\label{Language}
\begin{tabular}{|l|l|l|}
\hline
Language                                                                                      & Features                                                                                                                         & Reference                         \\ \hline
\begin{tabular}[c]{@{}l@{}}PDDL\\ (Planning Domain\\Description Language)\end{tabular}         & \begin{tabular}[c]{@{}l@{}}Machine-readable,\\ standardized syntax \\ for representing STRIPS\\ and other languages.\\Has types, constants,\\predicates and actions\end{tabular} & \cite{mcdermott1998pddl}        \\ \hline
\begin{tabular}[c]{@{}l@{}}STRIPS\\ (Stanford Research\\Institute Problem\\Solver)\end{tabular} & Sublanguage of PDDL                                                                                                              & \cite{fikes1971strips}          \\ \hline
\begin{tabular}[c]{@{}l@{}}OCL\\ (Object Centered\\Language)\end{tabular}                      & \begin{tabular}[c]{@{}l@{}}High level language\\ with representation centered\\ around objects instead \\ of states\end{tabular}                            & \cite{mccluskey2002interactive} \\ \hline
\begin{tabular}[c]{@{}l@{}}STRIPS+WS  \end{tabular}                                                                                   &\begin{tabular}[c]{@{}l@{}} STRIPS + functional terms,\\leading to\\ higher expressiveness \end{tabular}                                                                                                       & \cite{walsh2008efficient}       \\ \hline
\end{tabular}
\end{table}

\section{State of the Art}

Brief descriptions of some key algorithms corresponding to the aforementioned classification can be found in the following section. These algorithms are also summarized in the table~\ref{table}.

\subsection{OBSERVER}
OBSERVER \cite{wang1996learning} is a system that learns operator preconditions by creating and updating both most specific representation and a most general representation for the preconditions, based on operator executions while solving practice problems. It also learns operator effects by generalizing the delta-state (the difference between post-state and pre-state) from multiple observations.

%OBSERVER (\cite{wang1996learning}) is a system that integrates planning, learning, and execution. It's learning module learns the preconditions and effects of operators by observing expert solution traces and to further refine the operators through practice by solving problems in the simulator using a learning-by-doing mechanism. Learning the operator preconditions consists of creating and updating both most specific representation and a most general representation for the preconditions, based on operator executions while solving practice problems. It also learns operator effects by generalizing the delta-state (the difference between post-state and pre-state) from multiple observations.

\subsection{RIM}
%RIM (Refining Incomplete planning domain models through plan traces) is an algorithm proposed by Zhuo et al. (\cite{zhuo2013refining}), which begins with the collection of sets of predicates, action schemas and macro-operator schemas from incomplete action models  and plan traces . Subsequently, it constructs sets of soft and hard constraints to ensure that the learned domain model can best explain the input plan traces and incomplete action models. These constraints are designed to be soft, directing the MAX-SAT algorithm in the learning step towards learning the most probable complete description of actions and macro-operators. Then, a set of hard constraints is enforced that must be satisfied by action models and macro-operators. Finally, these constraints are solved using a weighted MAX-SAT solver to obtain sets of macro-operators and (refined) action models. Macro-operators are used in this context to increase the accuracy of the incomplete model.

RIM (Refining Incomplete planning domain Models through plan traces) \cite{zhuo2013refining} constructs sets of soft and hard constraints which are solved using a weighted MAX-SAT solver to obtain sets of macro-operators and (refined) action models.

\subsection{OpMaker}
Opmaker \cite{mccluskey2002interactive} is a mixed initiative (where both the human and the machine take initiative), graphical knowledge acquisition tool for inducing parametrized, hierarchical (each object may have relations and attributes inherited from different levels \cite{mccluskey2002interactive}) operator descriptions from example action sequences and declarative domain knowledge, with the minimum of user interaction. It is implemented inside of a graphic tool called GIPO (Graphical Interface for Planning with Objects), which facilitates domain knowledge capture and  domain modelling (\cite{mccluskey2002interactive,jilani2014automated}), a perfect tool for novice users to create plans and learn models with minimum effort.
%It uses a high level language called OCL (Object Centered Language) for domain modelling. As Opmaker creates an operator schema from each action in a training solution sequence, it asks the user to input, if needed, the target state that each object would occupy after execution of the action. The algorithm induces operators, using a high level, partial model of the domain, as well as using the causal structure of the example sequence to track the state of an object.

%        \begin{figure}[h]
%        \centering
%        \includegraphics[width=100mm]{images/opmakerInt}
%        \caption{Opmaker Screen}
%        \label{fig:opmaker}
%        \end{figure}

\subsection{LAMP}
%LAMP (Learning Action Models from Plan traces) proposed by Zhuo et al.(\cite{zhuo2010learning})
%learn action models with quantifiers and logical implications . Firstly, the input plan traces(including observed states and actions) are encoded into propositional formulas, which is a conjunction of ground literals to store into a database as a collection of facts. These formulas are very close to PDDL in terms of representation. Secondly, candidate formulas are generated according to the predicate lists and domain constraints. Thirdly, a Markov Logic Network (MLN) uses the formulas generated in the above two steps to learn the corresponding weights of formulas to select the most likely subset from the set of candidate formulas. Finally, this subset is converted into the final action models.
LAMP (Learning Action Models from Plan traces) \cite{zhuo2010learning} learn action models with quantifiers and logical implications. Firstly, the input plan traces (including observed states and actions) are encoded into propositional formulas, which is a conjunction of ground literals to store into a database as a collection of facts. Secondly, candidate formulas are generated according to the predicate lists and domain constraints. Thirdly, a Markov Logic Network (MLN) uses the formulas generated in the above two steps to select the most likely subset from the set of candidate formulas. Finally, this subset is converted into the final action models.

%\begin{figure}[h]
%    \centering
%    \includegraphics[width=90mm]{images/table1}
%    \caption{Planning and Learning (D:Deterministic, FO:Fully Observable, PO: Partially Observable)}
%    \label{fig:planningLearning}
%    \end{figure}

\subsection{AMAN}
%AMAN (Action-Model Acquisition from Noisy plan traces) \cite{zhuo2013action} finds a domain model that best explains the observed noisy plan traces. First, a set of candidate domain models is built by scanning each action in plan traces and substitute its instantiated parameters with their corresponding variables , thus building a set of action schemas. This is followed by a graphical model to capture the relationship between the current state, correct action, observed action and the domain model. Afterwards the parameters of the graphical model are learnt, following which AMAN generates a set of action models according to the learnt parameters. Specifically, first the observed noisy plan traces are exploited to predict correct plan traces and the domain model based on the graphical model. Then the correct plan traces are executed to calculate the reward of the predicted correct plan traces according to a predefined reward function.  The AMAN algorithm framework belongs to a family of policy gradient algorithms, which have been successfully applied to complex problems.

AMAN (Action-Model Acquisition from Noisy plan traces) \cite{zhuo2013action} finds a domain model that best explains the observed noisy plan traces. First, a set of candidate domain models is built by scanning each action in plan traces and substituting its instantiated parameters with their corresponding variables. This is followed by a graphical model to capture the relationship between the current state, correct action, observed action and the domain model. Afterwards the parameters of the graphical model are learnt, following which AMAN generates a set of action models according to the learnt parameters.
%Specifically, first the observed noisy plan traces are exploited to predict correct plan traces and the domain model based on the graphical model. Then the correct plan traces are executed to calculate the reward of the predicted correct plan traces according to a predefined reward function.

\subsection{EXPO}
%The EXPO (\cite{wang1996learning,jimenez2013integrating,gil1992acquiring}) system refined incomplete planning operators, that is, operators with some over-general preconditions and missing effects by means of ORM (operator refinement method). EXPO does this by generating plans and monitoring their execution to detect the differences between the state predicted according to the internal action model and the observed state. EXPO then constructs a set of specific hypotheses to fix the detected differences. After being heuristically filtered, each hypothesis is tested in turn with an experiment and a plan is constructed to achieve the situation required to carry out the experiment. This approach is also interesting because it also allows for the formation of micro operators in situations where only some effects of operator are required.

%A summary of the aforementioned algorithms can be found in the tables ~\ref{fig:planningLearning} and ~\ref{fig:planningLearning(II)} below.
The EXPO (\cite{wang1996learning,jimenez2013integrating,gil1992acquiring}) system refines incomplete planning operators by ORM (operator refinement method). EXPO does this by generating plans and monitoring their execution to detect the differences between the state predicted according to the internal action model and the observed state. EXPO then constructs a set of specific hypotheses to fix the detected differences. After being heuristically filtered, each hypothesis is tested in turn with an experiment and a plan is constructed to achieve the situation required to carry out the experiment.

\subsection{ARMS}
%The ARMS (Action-Relation Modelling System) \cite{yang2007learning} system learns the action model from traces obtained by an observation agent who does not know the logical encoding of the actions and the full state information between the actions. This action model is not guaranteed to be completely correct, but it can serve to provide important initial guidance for human knowledge editors. ARMS proceeds in two phases. In phase one of the algorithm, ARMS finds frequent action sets from plans that share a common set of parameters. In addition, ARMS finds some frequent relation-action pairs with the help of the initial state and the goal state. These relation-action pairs give us an initial guess on the preconditions, add lists and delete lists of actions in this subset. These action subsets and pairs are used to obtain a set of constraints that must hold in order to make the plans correct. The constraints extracted from the plans are then transformed into a weighted MAX-SAT representation, the solution to which produces action models. The process iterates until all actions are modelled.

The ARMS (Action-Relation Modelling System) \cite{yang2007learning} system learns an action model in two phases. In phase one of the algorithm, ARMS finds frequent action sets from plans that share a common set of parameters. In addition, ARMS finds some frequent relation-action pairs with the help of the initial state and the goal state. These relation-action pairs give us an initial guess on the preconditions, add lists and delete lists of actions in this subset. These action subsets and pairs are used to obtain a set of constraints that must hold in order to make the plans correct. The constraints extracted from the plans are then transformed into a weighted MAX-SAT representation, the solution to which produces action models. The process iterates until all actions are modelled.

\subsection{PELA}
%PELA (Planning, Execution and Learning Architecture)\cite{jimenez2008architecture}, as its name suggests, performs three principal functions: (i) Planning the actions that solve a given problem (2) Execution of plans and classification of the execution outcomes. PELA executes plans one action at a time in the environment and labels the actions executions according to their outcomes as success, failure or dead-end. The learning component allows PELA to generate probabilistic rules about the execution of actions. PELA generates these rules from the execution of plans and compiles them to upgrade its deterministic planning model. This is done by performing multiclass classification, which further consists of finding the smallest decision tree that fits a given data set. The common way to find these decision trees is following a Top-Down Induction of Decision Trees (TDIDT) algorithm \cite{quinlan1986induction}. This approach builds the decision tree by splitting the learning examples according to the values of a selected attribute that minimize a measure of variance along the prediction variable.

PELA (Planning, Execution and Learning Architecture) \cite{jimenez2008architecture} performs the three functions suggested in its name. The learning component allows PELA to generate probabilistic rules about the execution of actions. PELA generates these rules from the execution of plans and compiles them to upgrade its deterministic planning model. This is done by performing multiclass classification, which further consists of finding the smallest decision tree that fits a given data set following a Top-Down Induction of Decision Trees (TDIDT) algorithm \cite{quinlan1986induction}.% This approach builds the decision tree by splitting the learning examples according to the values of a selected attribute that minimize a measure of variance along the prediction variable.

\subsection{LAWS}
%One particular implementation is an algorithm called LAWS (Learn Action models with transferring knowledge from a related source domain via Web search) \cite{zhuo2011cross}. Given just a limited amount of training data, it make use of action-models already created beforehand in other related domains, which are called source domains, to help learn actions in a target domain. The target domain and a related source domain are bridged by searching Web pages related to the target domain and the source domain, and then building a mapping between them by means of a similarity function done by calculating the similarity between their corresponding Web pages.  The similarity is calculated using the Kullback-Leibler (KL) divergence or Maximum Mean Discrepancy (MMD). Based on the calculated similarity, a set of weighted constraints, called web constraints, are built. Based any available example plan traces in the target domain, other constraints such as state constraints, action constraints and plan constraints, are also built. All the constraints (web/state/action/plan constraints) are solved using a weighted MAX-SAT solver, and target-domain action models are generated based on the solution to the constraint satisfaction problem.

LAWS (Learn Action models with transferring knowledge from a related source domain via Web search) \cite{zhuo2011cross} makes use of action-models already created beforehand in other related domains, which are called source domains, to help learn actions in a target domain. The target domain and a related source domain are bridged by searching Web pages related to the target domain and the source domain, and then building a mapping between them by means of a similarity function done by calculating the similarity between their corresponding Web pages.  The similarity is calculated using the Kullback-Leibler (KL) divergence. Based on the calculated similarity, a set of weighted constraints, called web constraints, are built. Based any available example plan traces in the target domain, other constraints such as state constraints, action constraints and plan constraints, are also built. All the above constraints are solved using a weighted MAX-SAT solver, and target-domain action models are generated based on the solution to the constraint satisfaction problem.

\subsection{LOPE}
%LOPE (Learning by Observation in Planning Environments) \cite{garcia2000integrated} speaks of learning by sharing among multi agent systems. Each agent continuously learns, plans and executes. Each of the agents receives as input: perceptions from the environment (situations and utilities); set of actions that it can perform; and operators learned by other agents. The output of each agent is a sequence of actions over time (for the environment), and, regularly, the set of operators that it learned. An agent either shares all its operators with other agents, or its most reliable ones.  Learning is performed by three integrated techniques: rote learning of an experience (observation) by creating an operator directly from it; heuristic generalization of incorrect learned operators; and a global reinforcement strategy of operators by rewarding and punishing them based on their success in predicting the behavior of the environment. Reinforcement of an operator means punishment of similar ones, so there is a global reinforcement of the same action. This global reinforcement is done by means of a virtual generalized Q table \cite{garcia2000integrated}.
LOPE (Learning by Observation in Planning Environments) \cite{garcia2000integrated} learns by sharing among multi agent systems. Learning is performed by three integrated techniques: rote learning of an experience (observation) by creating an operator directly from it, heuristic generalization of incorrect learned operators; and a global reinforcement strategy of operators by rewarding and punishing them based on their success in predicting the behavior of the environment. Reinforcement of an operator means punishment of similar ones, so there is a global reinforcement of the same action. This global reinforcement is done by means of a virtual generalized Q table \cite{garcia2000integrated}.

\subsection{FOOLMETWICE}
%FOOLMETWICE \cite{molineaux2014learning} is a goal-oriented (GDA-Goal \\Driven Agent) algorithm which learns from surprises. It tries to find inaccuracies in environment model $M$ by attempting to explain all observations received. When a consistent explanation cannot be found, it infers that some unknown event $E$ happened that is not represented in $M$. However, it operates in a partially observable environment; events and their effects are not always immediately observed. Before inferring a model for $E$, it must determine when $E$ most likely occurred. It does this by finding a minimally inconsistent explanation I that is more plausible than any other such explanation that can be described based on the current model and observations. $E$ does not include any specific unknown event, but does help to pinpoint when it occurred. To search for minimally inconsistent explanations, the algorithm DISCOVERHISTORY \cite{molineaux2011learning} has been extended to ignore an inconsistency by creating an inconsistency patch. If an explanation describes all correct events, unknown events correspond to inconsistency patches; the unknown effects are the same as those of the patch events. After determining when unknown events occur, creating a model of their preconditions requires generalizing over the states that trigger them. Instead of inferring rules from a relational database, this implementation infers rules from a set of projected states, each of which contains all facts believed to be true at a specific prior time.\\
FOOLMETWICE \cite{molineaux2014learning} is a goal-oriented (GDA-Goal Driven Agent) algorithm which learns from surprises. It tries to find inaccuracies in environment model $M$ by attempting to explain all observations received. When a consistent explanation cannot be found, it infers that some unknown event $E$ happened that is not represented in $M$. After determining when unknown events occur, it creates a model of their preconditions requires generalizing over the states that trigger them.

\section{Open Issues}
Despite the bright prospects that the aforementioned approaches offer, there persist some open issues and loopholes which are discussed as follows:
\begin{itemize}
\item Learning with the time dimension: Time plays an imperative role most real life domains. For example, each dialogue in a HRI is composed of an utterance further accompanied by gestural, body and eye movements, all of them interleaved in a narrow time frame. These interactions may thus be represented by a time sequence, with the intent of learning the underlying action model. Barring some initial works in this area, time remains an interesting aspect to explore \cite{zhang2015capability}.

\item Direct re-applicability of learned model for dialogue exchange: re-use of a learned model by a planner continues to remain a concern. A model that has been learned by applying ML techniques is more often than not incomplete, or more concretely inadept to be fed to a planner to directly generate plans, or in the case of HRI, to reproduce a multimodal dialogue which respects social rules. It needs to be retouched and fine tuned by a domain expert in order to be reusable. This marks a stark incapability of the prominently used machine learning techniques to be and comprehensive, and leaves scope for much more research.

\item Extension of classical planning to a full scope domain: The applicability of the aforementioned approaches, most of which have been tested on highly simplified toy domains and not in real scenarios, remains an issue to be addressed. Classical planning refers to a world model in which predicates are propositional: they do not change unless the planning agent acts to change them, all relevant attributes can be observed at any time, the impact of executing an action on the environment is known and deterministic, the effects of taking an action occurring instantly and so on. However, the real world is laced with unpredictability: a predicate might switch its value spontaneously, the world may have hidden variables, the exact impact of actions may be unpredictable, the actions may have durations and so on \cite{zimmerman2003learning}.
\end{itemize}

\section{Conclusion}

This article argues for the usage of AI planning techniques with the intent of endowing robots with socio-communicative skills, thus augmenting their acceptability. It justifies the notion of learning the underlying behavioral blueprint of the robot from a set of multimodal HRI traces. This learning is achieved by the usage of several state-of-the-art and classical Machine Learning (ML) techniques. The article tries to classify various ML approaches based on several criterion and conditions, along with the merits and demerits of each approach. It then broadly highlights some persisting open issues with the discussed approaches, concluding that a significant number of prominent and interesting techniques have been applied to highly controlled experimental setups, and their application to a real world HRI scenario is a topic of further research.

\begin{landscape}
\setlength{\oddsidemargin}{-10pt} % Marge gauche sur pages impaires
 \begin{table*}[]
\tiny
\centering
\caption{State Of The Art Planning Algorithms (D=Deterministic, P=Probabilistic, FO=Fully Observable, PO=Partially Observable, PDL=Prodigy Description Language)}
\label{table}
\begin{tabular}{|l|l|l|l|l|l|l|l|l|}
\hline
\begin{tabular}[c]{@{}l@{}}Algorithm/\\ Author\end{tabular}                    & Input                                                                                         & Output                            & Language                                               & Technique& Merits                                                                                                                                                                                                            & Demerits                                                                                                                                                                                                                                        & \begin{tabular}[c]{@{}l@{}}Envi\\ro\\nment \end{tabular}& \begin{tabular}[c]{@{}l@{}}Robust \\ to\\ Noise?\end{tabular} \\ \hline
\begin{tabular}[c]{@{}l@{}}OBSE\\RVER\\ (\cite{wang1996learning})\end{tabular} & \begin{tabular}[c]{@{}l@{}}Traces,\\practice\\problems and\\description\\language\end{tabular} & Operators                          & \begin{tabular}[c]{@{}l@{}}STRIPS-\\ like\end{tabular} & \begin{tabular}[c]{@{}l@{}}Conservative\\ specific-to-\\general \\inductive \\ generalization\\ process\end{tabular}& \begin{tabular}[c]{@{}l@{}}(i) Can find out negated \\preconditions,\\ conditional preconditions\\ and \\conditional effects\\(ii) Does not require \\strong background\\ knowledge\end{tabular}                                    & \begin{tabular}[c]{@{}l@{}}Domain knowledge can be \\incomplete or incorrect\\  in the following ways : \\over-general preconditions,\\  over-specific preconditions,\\ incomplete effects,\\  extraneous effects, \\and missing operators\end{tabular} & \begin{tabular}[c]{@{}l@{}}D,\\ FO\end{tabular}       & N                                                           \\ \hline

\begin{tabular}[c]{@{}l@{}}RIM\\(\cite{zhuo2013refining})\end{tabular} & \begin{tabular}[c]{@{}l@{}} Incomplete\\action \\models \\and plan\\traces \end{tabular}& \begin{tabular}[c]{@{}l@{}}Macro-operators \\and action models\end{tabular} & STRIPS & MAX-SAT & \begin{tabular}[c]{@{}l@{}}Increases accuracy of\\ incomplete operators\end{tabular} & \begin{tabular}[c]{@{}l@{}}Cannot handle incorrect \\partial initial models\end{tabular} & \begin{tabular}[c]{@{}l@{}}D,\\ FO\end{tabular}& Y \\
\hline

\begin{tabular}[c]{@{}l@{}}Opmaker\\(\cite{mccluskey2002interactive}) \end{tabular}& \begin{tabular}[c]{@{}l@{}}Partial Model, \\Action\\sequences\end{tabular} & \begin{tabular}[c]{@{}l@{}}Operator Schema \end{tabular}& OCL & \begin{tabular}[c]{@{}l@{}}Operator\\ induction by\\ mixed \\initiative\end{tabular} &\begin{tabular}[c]{@{}l@{}} (i) Eases task of \\operator encoding, fits well \\into engineering environment\\ for planning domain \\acquisition\\ and modeling \\(ii) Useful for non-experts\end{tabular} &\begin{tabular}[c]{@{}l@{}}Needs user input\\ for intermediate state \\information\end{tabular} & \begin{tabular}[c]{@{}l@{}}D,\\ FO\end{tabular} & Y \\
\hline

\begin{tabular}[c]{@{}l@{}}LAMP\\ (\cite{zhuo2010learning})\end{tabular} & \begin{tabular}[c]{@{}l@{}}State-Action\\Interleavings \end{tabular}& \begin{tabular}[c]{@{}l@{}}Action Models\end{tabular} & PDDL & \begin{tabular}[c]{@{}l@{}}Markov\\Logic\\Network (MLN)\end{tabular}& \begin{tabular}[c]{@{}l@{}}More expressive models-\\inculcate quantifiers\\and logical implications\end{tabular}&\begin{tabular}[c]{@{}l@{}} Looses efficiency\\with increasing\\domain size \end{tabular}&\begin{tabular}[c]{@{}l@{}} D,\\ FO\end{tabular} & N \\ \hline

\begin{tabular}[c]{@{}l@{}}AMAN\\ (\cite{zhuo2013action})\end{tabular} &\begin{tabular}[c]{@{}l@{}} Action\\ sequence\end{tabular} & Operators & STRIPS & \begin{tabular}[c]{@{}l@{}}Gradient\\ Descent, \\Reinforcement\\ Learning\end{tabular} & \begin{tabular}[c]{@{}l@{}}No background knowledge \\needed \end{tabular}& \begin{tabular}[c]{@{}l@{}}Model sampling \\mechanism \\unclear\end{tabular} & \begin{tabular}[c]{@{}l@{}}D,\\ FO\end{tabular} & Y \\ \hline

\begin{tabular}[c]{@{}l@{}}EXPO\\ (\cite{wang1996learning})\end{tabular} & \begin{tabular}[c]{@{}l@{}}Incomplete\\operator\\ set,\\ traces of state\\ sequences \end{tabular}& \begin{tabular}[c]{@{}l@{}}New preconditions,\\ effects,\\ conditional effects,\\operators, attribute\\ values\end{tabular} & \begin{tabular}[c]{@{}l@{}} PDL \end{tabular}& \begin{tabular}[c]{@{}l@{}}Learning-by-\\experimentation \\for\\ Operator \\Refinement \end{tabular}&\begin{tabular}[c]{@{}l@{}} (i) Learns conditional effects\\(ii) Methods are goal-directed \\and learning is incremental\end{tabular} &\begin{tabular}[c]{@{}l@{}} Rules learnt from general\\ to specific\end{tabular} & \begin{tabular}[c]{@{}l@{}}D,\\ FO\end{tabular} & N \\
\hline

\begin{tabular}[c]{@{}l@{}}ARMS\\ (\cite{yang2007learning})\end{tabular} & \begin{tabular}[c]{@{}l@{}}Action\\sequence\\ with partial\\traces \end{tabular}&\begin{tabular}[c]{@{}l@{}} Action Models \end{tabular}& STRIPS &\begin{tabular}[c]{@{}l@{}} Builds a weighted \\propositional\\ satisfiability problem\\ and solves it using\\ weighted MAX-SAT\\ solver\end{tabular} & \begin{tabular}[c]{@{}l@{}}Can handle cases when\\ intermediate state\\ observations\\ are difficult to acquire\end{tabular} & \begin{tabular}[c]{@{}l@{}}(i) Cannot learn action\\ models\\ with quantifiers \\or implications\\ (ii) Cannot learn \\complex\\ action models\end{tabular} & \begin{tabular}[c]{@{}l@{}}D,\\ PO\end{tabular} & N \\
\hline

\begin{tabular}[c]{@{}l@{}}PELA\\ (\cite{jimenez2008architecture})\end{tabular} &\begin{tabular}[c]{@{}l@{}} Planning\\problem,\\STRIPS\\action\\model \end{tabular}&\begin{tabular}[c]{@{}l@{}} Enriched action\\ model with \\planning\\ possibilities in\\ probabilistic \\domains\end{tabular} & PDDL & \begin{tabular}[c]{@{}l@{}}TDIDT (Top Down\\ Induction\\ of Decision Trees) \end{tabular}&\begin{tabular}[c]{@{}l@{}} Based on off-the-shelf \\planning and \\learning components\end{tabular} & \begin{tabular}[c]{@{}l@{}}Assumes initial action\\ model of environment \end{tabular} & \begin{tabular}[c]{@{}l@{}}P,\\ FO\end{tabular}& N \\ \hline

\begin{tabular}[c]{@{}l@{}}LAWS\\(\cite{zhuo2011cross}) \end{tabular}&\begin{tabular}[c]{@{}l@{}} Action\\schemas,\\ predicates,\\ plan traces\\ from target\\domain,\\action models\\ from source\\domain\end{tabular} &\begin{tabular}[c]{@{}l@{}} Action models\\ in target domain \end{tabular}& STRIPS &\begin{tabular}[c]{@{}l@{}}Transfer Learning\\,KL divergence \end{tabular} &\begin{tabular}[c]{@{}l@{}} Web exploited \\as knowledge source \end{tabular}&\begin{tabular}[c]{@{}l@{}} Negative Transfer:\\ When source domain\\ and target\\domain are\\ not related to each other,\\ brute-force transfer\\ may be unsuccessful\end{tabular} & \begin{tabular}[c]{@{}l@{}}D,\\ FO\end{tabular} & N \\ \hline

\begin{tabular}[c]{@{}l@{}}LOPE\\ (\cite{garcia2000integrated})\end{tabular} & \begin{tabular}[c]{@{}l@{}}Number of\\execution\\cycles,\\possible\\action set\end{tabular} & Operators & \begin{tabular}[c]{@{}l@{}}Propos\\itional\\ Logic \end{tabular}&\begin{tabular}[c]{@{}l@{}} Reinforcement \\Learning \end{tabular}&\begin{tabular}[c]{@{}l@{}} Allows knowledge sharing\\among agents,\\increasing percentage\\of successful plans \end{tabular}&\begin{tabular}[c]{@{}l@{}} Differences in sensors\\of agents causes\\different ways of\\perceiving the world,\\and, therefore, different\\biases towards\\operator generation\end{tabular} &\begin{tabular}[c]{@{}l@{}} D,\\FO \end{tabular}& N \\ \hline

\begin{tabular}[c]{@{}l@{}}FOOL\\ME\\TWICE\\ (\cite{molineaux2014learning}) \end{tabular}&\begin{tabular}[c]{@{}l@{}} Initial state,\\ goal state,\\ model \end{tabular}&\begin{tabular}[c]{@{}l@{}} Expectation,\\ discrepancies,\\ goals \end{tabular}& PDDL+ &\begin{tabular}[c]{@{}l@{}} Goal-oriented \\model\\ based on surprise\\ and explanation\\ generation\end{tabular}  &\begin{tabular}[c]{@{}l@{}} Goal-oriented rather\\ than reward-driven, \\thus allowing frequent\\ goal change without\\ requiring substantial policy\\ re-learning\end{tabular} &\begin{tabular}[c]{@{}l@{}} Cannot acquire exogenous\\ event models with\\ continuous conditions \end{tabular} & \begin{tabular}[c]{@{}l@{}}D,\\ PO\end{tabular} & N \\ \hline

\end{tabular}
\end{table*}
\end{landscape}

\end{document}